\documentclass[runningheads]{llncs}

\usepackage{eccv}

\usepackage{eccvabbrv}

\usepackage{graphicx}
\usepackage{booktabs}
\usepackage{multirow}
\usepackage{lipsum}

\usepackage[accsupp]{axessibility}  % Improves PDF readability for those with disabilities.

\usepackage{hyperref}

\usepackage{orcidlink}

\newcommand\blfootnote[1]{%
\begingroup
\renewcommand\thefootnote{}\footnote{#1}%
\addtocounter{footnote}{-1}%
\endgroup
}

\begin{document}

% ---------------------------------------------------------------
% TODO REVIEW: Replace with your title
\title{Leveraging Text Localization for Scene Text Removal via Text-aware Masked Image Modeling} 

% TODO REVIEW: If the paper title is too long for the running head, you can set
% an abbreviated paper title here. If not, comment out.
\titlerunning{TMIM}

% TODO FINAL: Replace with your author list. 
% Include the authors' OCRID for the camera-ready version, if at all possible.
\author{Zixiao Wang\inst{1}\orcidlink{0000-0002-0009-5033} \and
Hongtao Xie\inst{1}\textsuperscript{$\star$}\orcidlink{0000-0002-6249-5315} \and
YuXin Wang\inst{1}\orcidlink{0000-0002-0228-6220} \and
Yadong Qu\inst{1}\orcidlink{0000-0003-0265-5011} \and
Fengjun Guo\inst{2} \and
Pengwei Liu\inst{2} } 

% TODO FINAL: Replace with an abbreviated list of authors.
\authorrunning{Wang et al.}
% First names are abbreviated in the running head.
% If there are more than two authors, 'et al.' is used.

% TODO FINAL: Replace with your institution list.
\institute{University of Science and Technology of China \and
IntSig Information Co., Ltd \\
\email{\{wzx99,wangyx58,qqqyd\}@mail.ustc.edu.cn, htxie@ustc.edu.cn} \\
\email{\{fengjun\_guo,pengwei\_liu\}@intsig.net}\\
\blfootnote{\textsuperscript{$\star$}Corresponding Author}
}

\maketitle

\begin{abstract}
    Existing scene text removal (STR) task suffers from insufficient training data due to the expensive pixel-level labeling.
    In this paper, we aim to address this issue by introducing a Text-aware Masked Image Modeling algorithm (TMIM), which can pretrain STR models with low-cost text detection labels (e.g., text bounding box).
    Different from previous pretraining methods that use indirect auxiliary tasks only to enhance the implicit feature extraction ability, our TMIM first enables the STR task to be directly trained in a weakly supervised manner, which explores the STR knowledge explicitly and efficiently.
    In TMIM, first, a Background Modeling stream is built to learn background generation rules by recovering the masked non-text region. 
    Meanwhile, it provides pseudo STR labels on the masked text region.
    % It then serves as the pseudo labels provider for the unlabeled STR images. 
    Second, a Text Erasing stream is proposed to learn from the pseudo labels and equip the model with end-to-end STR ability. Benefiting from the two collaborative streams, our STR model can achieve impressive performance only with the public text detection datasets, which greatly alleviates the limitation of the high-cost STR labels.
    Experiments demonstrate that our method outperforms other pretrain methods and achieves state-of-the-art performance (37.35 PSNR on SCUT-EnsText).
    Code will be available at https://github.com/wzx99/TMIM.
\end{abstract}

\section{Introduction}
\label{sec:intro}

\begin{figure}[t]
  \centering
   \includegraphics[width=0.9\linewidth]{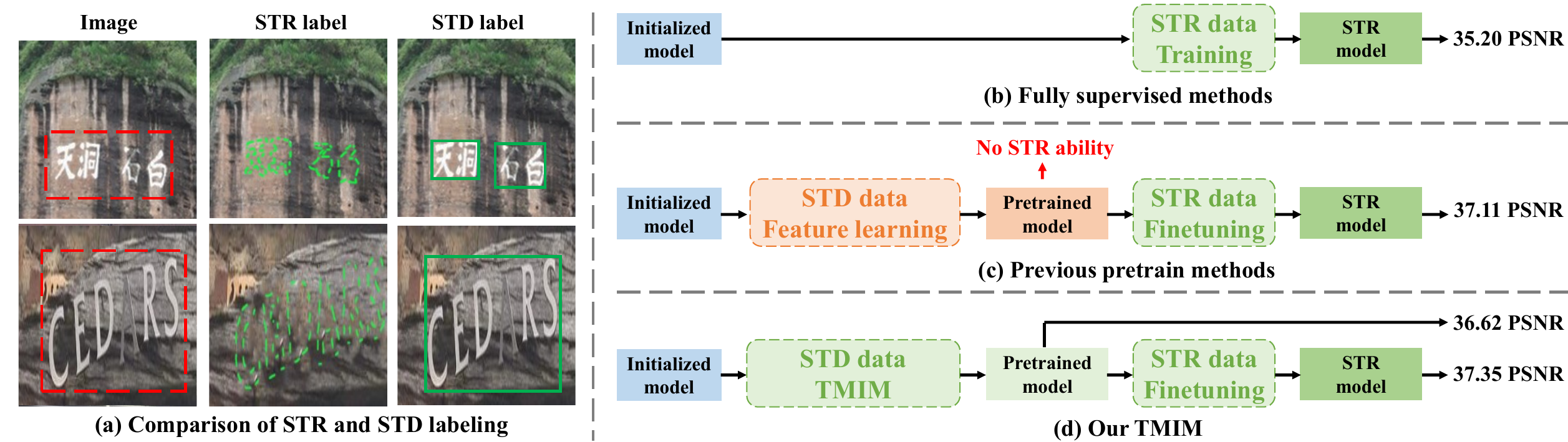}
   \caption{(a)Comparison of the annotation in scene text removal (STR) task and scene text detection (STD) task. It is clear that STR requires fine-grained manual modification. While STD only requires box-level text detection labeling which can be obtained by advanced OCR systems. (b-c) previous fully supervised \cite{liu2022don} and pretraining \cite{peng2023viteraser} methods. (d)our TMIM.}
   \label{fig:img_diff}
\end{figure}

Scene text removal (STR) plays an important role in many real-world applications, such as privacy protection and picture editing \cite{wang2023real,liu2020erasenet,zhang2019ensnet}.
It aims to delete the text regions in the given image and fill them with proper background content.
Recently, with the success of the EnsNet \cite{zhang2019ensnet}, deep learning based methods have been widely applied in STR tasks \cite{liu2020erasenet,tursun2019mtrnet,tang2021stroke}.
These data-driven algorithms demonstrate promising text removal ability under the supervision of pixel-level annotations. However, compared with labels in other fundamental OCR tasks such as scene text detection (STD) task (\cref{fig:img_diff}(a)), annotating a large number of pixel-level text removal labels is expensive and time-consuming (5 - 10 minutes per image \cite{liu2020erasenet}). As a result, the size of annotated STR datasets is relatively small (e.g., 2.7k in \cite{liu2020erasenet} vs 111k in existing STD datasets as explained in \cref{sec:dataset}), which is inadequate to fully exploit the potential of advanced STR models.

Aiming at addressing the issue of limited training data, some previous works utilize synthetic datasets for training \cite{gupta2016synthetic,zhang2019ensnet}. 
To enhance the quality of synthetic data, Tang et al. \cite{tang2021stroke} and Jiang et al. \cite{jiang2022self} develop the Text Synthesis Engine to synthesize more challenging texts.
But the domain gap between synthetic and real data can not be perfectly eliminated, resulting in limited performance when applied to real data.
Other works propose to pretrain models on additional data.
Zdenek et al. \cite{zdenek2020erasing} follow image inpainting approaches \cite{pathak2016context} and pretrain the model on general inpainting datasets.
To fit the data distribution of scene text images, Peng et al. \cite{peng2023viteraser} further introduce STD datasets for feature learning. 
They combine masked image modeling (MIM) and STD as pretraining tasks to enhance the feature representation.
However, when the texts are masked, the MIM pretraining tends to recover the masked text rather than remove it, which is in conflict with the STR target.
Thus, although the STD datasets can provide additional scene text images at a low cost, the lack of STR labels forces existing methods to design indirect pretraining tasks for implicit feature learning that lacks specificity to STR (\cref{fig:img_diff}(c)), resulting in redundancy and inefficiency.

To further enhance the STR pretraining efficiency, we focus on addressing the conflict between MIM and STR and propose a Text-aware Masked Image Modeling (TMIM) pretraining framework, which can directly employ STR training on STD datasets through a novel utilization of text detection labels.
The TMIM consists of two parallel streams: a Background Modeling (BM) stream for learning non-text background generation rules and a Text Erasing (TE) stream for end-to-end STR learning.
Specifically, in the BM stream, we first put masks on all text regions and random non-text regions, and then train the model to recover them.
In contrast to general MIM methods that aim to reconstruct all masked regions, we solely supervise the results on non-text regions.
Thus, the model is forced to concentrate on generating non-text content, 
which greatly eliminates the harmful ambiguity of whether to generate text or background in the pretraining stage.
Meanwhile, the prediction results on masked text regions can further serve as pseudo labels for STR learning.
Hence, we present the TE stream to utilize these labels for end-to-end text removal training, which explicitly enhances the STR capability.
Besides, we propose a Prompt-based Multi-task Learning strategy to enable the two streams to share the same model weights and train the two streams simultaneously, which efficiently transfers background generation knowledge in the BM stream to the TE stream for STR, thereby simplifying the training process. 
Compared with previous methods in \cref{fig:img_diff}(c-d), our TMIM first enables end-to-end STR training with only text detection labels, thereby alleviating the constraints imposed by the high-cost small-scale STR dataset.

It is worth noting that we focus on a data perspective for low-labeling-cost pretraining rather than designing a specific model architecture.
Since there already exists many large-scale labeled STD datasets(111k in STD vs 2.7k in STR as listed in \cref{sec:dataset}), we believe it is valuable for our TMIM to further explore the potential in them to alleviate the data limitation problem.
The proposed TMIM offers a straightforward and effective framework that can be simply applied to existing STR model architectures for stability performance improvement.
We investigate the results of our method on 5 advanced models, covering both CNN and Transformer architectures, which results in an over 0.34 PSNR improvement across all models.
Especially, our TMIM enhances Uformer-B \cite{wang2022uformer} to a new state-of-the-art PSNR (37.35) on the SCUT-EnsText dataset, surpassing other pretraining methods by 0.24.
Moreover, our weakly supervised pretrained models with only STD labels also achieve remarkable STR performance with 36.62 PSNR, outperforming all compared fully supervised no-finetuning models with 0.3 improvements.
These results validate that our TMIM effectively captures valuable STR knowledge from low-cost STD datasets. 
Overall, our approach presents a novel solution for label-limited STR training, which can serve as a plug-and-play method to support complex STR models from a new data perspective.
The contributions of our method are summarized as follows: 

\begin{itemize}
    \item We demonstrate that the STR ability can be obtained using text detection labels, which significantly alleviates the limitation of insufficient STR annotations and first enables STD datasets to train end-to-end STR models.
    \item We propose a novel weakly supervised Text-aware Masked Image Modeling framework (TMIM) for STR pretraining. By applying non-text generation rules on text regions, TMIM can transfer background modeling knowledge for text removal efficiently. 
    \item Our method outperforms other pretrain methods and achieves state-of-the-art performance on the SCUT-EnsText dataset (37.35 PSNR).
    And our pretrained model with only STD dataset also surpasses most existing STR methods (36.62 PSNR).
\end{itemize}

\section{Related Work}

\noindent\textbf{Scene text removal.} STR is an important technique for many applications and has become a research hotspot.
In the early years, traditional text removal methods \cite{khodadadi2012text,wagh2015text} follow a two-stage method that first detected the text localization and then use the manually designed inpainting algorithm to erase the texts.
With the development of deep learning, researchers tend to introduce deep neural networks for end-to-end text removal.
EnsNet \cite{zhang2019ensnet} combines a fully convolutional network and an adversarial network to reduce the artifacts.
As text region often takes up a small part of the whole image, some works \cite{liu2020erasenet,wang2023real,tursun2020mtrnet++,liu2022don,lee2022surprisingly,du2023modeling} propose to add a text region mask to help the network focus on text region.
For example, MTRNet \cite{tursun2019mtrnet} and MTRNet++ \cite{tursun2020mtrnet++} introduce an additional mask to guide the network to erase texts in the masked region.
Erasenet \cite{liu2020erasenet} adds text detection as an auxiliary task to enhance the perception of text regions.
PERT \cite{wang2023real} combines text detection and text removal results together and designs a region-based modification strategy to explicitly prevent erasing on non-text regions.
However, the data-driven deep learning algorithms heavily rely on training labels \cite{yang2022survey,ge2021semantic,ge2022dual}, which is insufficient in existing small-scale STR datasets.

To address this problem, Jiang et al. \cite{jiang2022self} propose a self-supervised synthesis algorithm to obtain high-quality synthetic data. 
Since synthetic data still has a domain gap with real data,
ViTEraser \cite{peng2023viteraser} introduces extra real text detection datasets and presents indirect self-supervised pretraining tasks for implicit representation learning.
Compared with them, this paper aims at using text detection data to directly train the model for explicit STR learning, which significantly enhances the training efficiency.

\noindent\textbf{Masked Image Modeling.} 
MIM aims to predict the masked region of the given image, which is a general pretraining method for various visual tasks.
BEiT \cite{bao2021beit} proposes to randomly replace some visual tokens with the mask token.
It trains the model to recover both the masked visual tokens and the original image.
MAE \cite{he2022masked} discards the masked tokens and trains the model to only focus on recovering the image.
SimMIM \cite{xie2022simmim} further uses larger masks and a lighter decoder to enhance the feature learning on the encoder.
Besides, SparK \cite{tian2023designing} entends the MIM framework for convolutional networks.
Recently, MIM-based pretraining methods have been widely applied to text analysis and understanding fields, such as MTM \cite{wang2023masked} for text detection, MARec \cite{jiang2023revisiting} for text recognition, and SegMIM \cite{peng2023viteraser} for text removal.
In contrast to them, we further develop MIM with a weakly supervised framework for explicit STR learning rather than implicit feature learning, which greatly enhances the pretraining efficiency.

\begin{figure*}[!t]
\centering
    \includegraphics[width=0.95\textwidth]{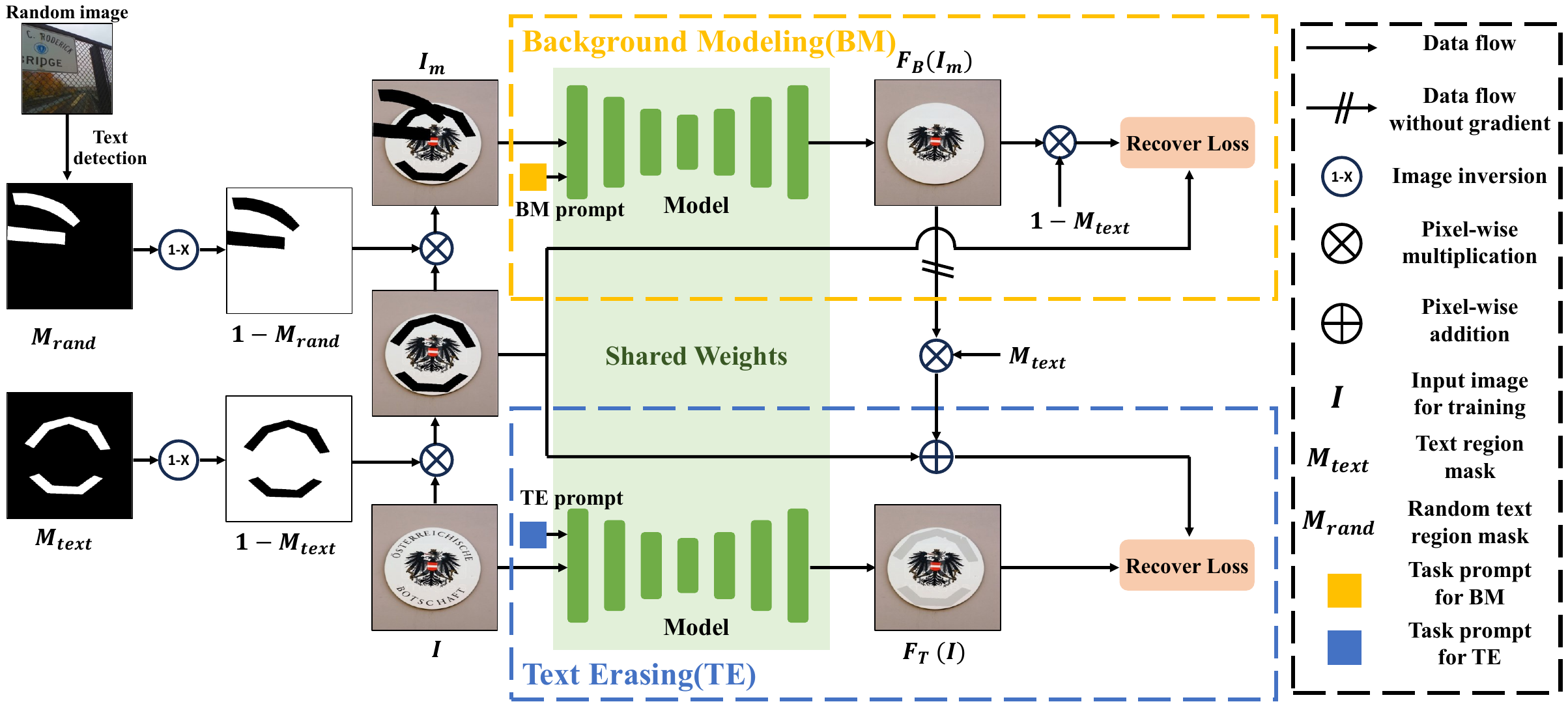}
    \caption{The overall framework of our TMIM, which consists of two streams: Background Modeling (BM) and Text Erasing (TE)stream. First, the BM stream uses the masked image as the input and trains the model to recover the background regions. Meanwhile, the TE stream adopts the recovery results from the BM stream to build the pseudo labels and train the model for STR.}
    \label{fig:img_framework}
\end{figure*}

\section{Methodology}
\label{sec:method}

\subsection{Pipeline}
For the pretraining stage, our TMIM employs weakly supervised STR learning on scene text images with text detection labels, which can be easily obtained from existing large-scale STD datasets or using an OCR system for automatic labeling.
The framework of TMIM is illustrated in \cref{fig:img_framework}.
The TMIM contains two parallel streams: Background Modeling (BM) stream and Text Erasing (TE) stream.
In the BM stream, we train the model to recover the masked regions with non-text content, granting the model background generation ability.
In the TE stream, we first combine the input image and the background predictions from the BM stream to obtain the pseudo STR label.
Then, we send the original input image into the model and utilize the pseudo label for end-to-end STR training.
During the optimization, the two streams are trained simultaneously and share the same model weights.
To enhance the stability of training two streams in a single model, we further introduce a Prompt-based Multi-task Learning strategy, which injects different task prompts into the model to handle each stream respectively.

For the finetuning stage, as the TE stream follows the general end-to-end STR training framework, we remove the BM stream and solely train the TE stream using the annotated STR dataset.
Besides, benefiting from the weakly supervised STR learning in the TE stream, our pretrained model can also achieve satisfactory STR performance without finetuning.

\subsection{Background Modeling stream}
\begin{figure*}[!t]
\centering
    \includegraphics[width=0.8\textwidth]{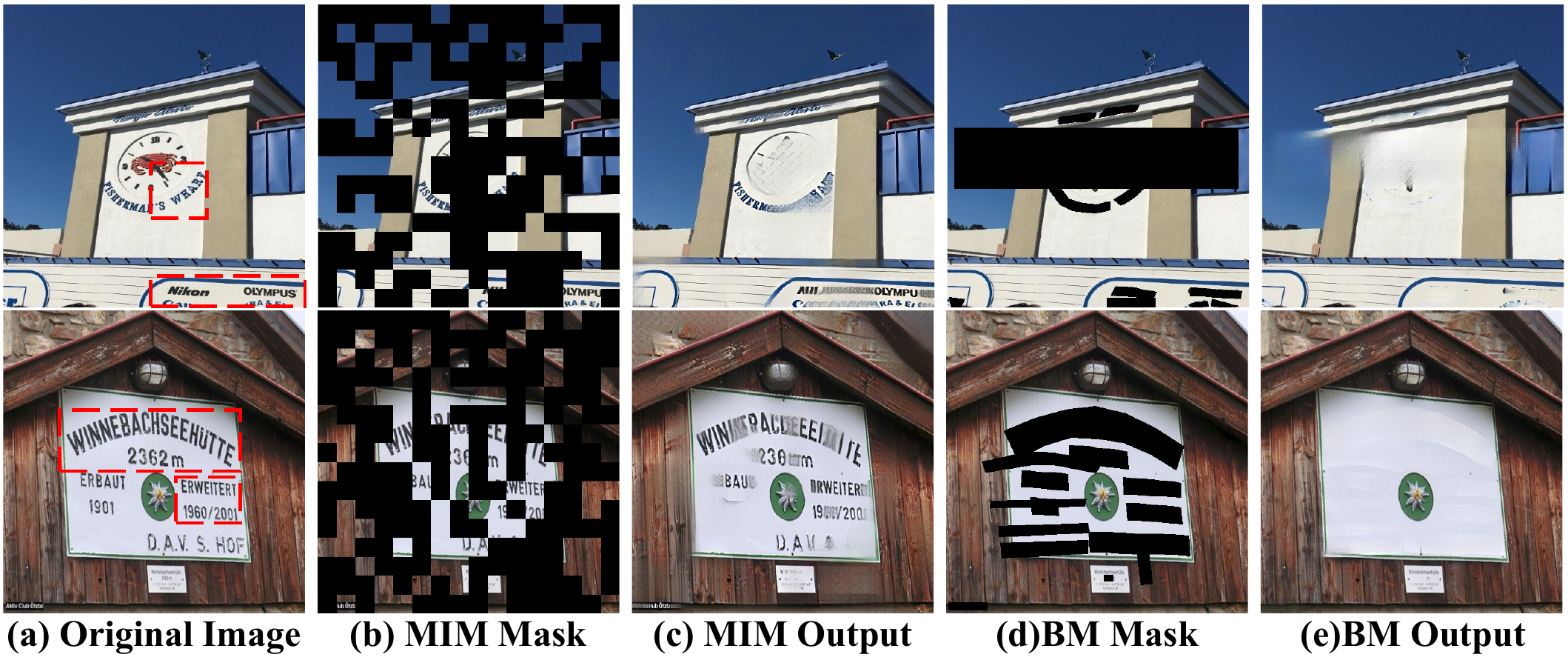}
    \caption{Comparison with MIM and our BM during the training process.  MIM tends to generate text-like content and our BM learns to erase texts.}
    \label{fig:img_inpaint}
\end{figure*}

Existing MIM methods typically apply a random mask to the input image, leading the model to reconstruct the masked regions and thereby learning for content generalization.
However, in STR, they tend to predict unfavorable text-like textures on the masked text regions (\cref{fig:img_inpaint}), which learns the opposite knowledge for text removal.
We believe this problem is caused by MIM trying to recover all contents in the masked regions which may also contain texts in it.
Thus, such general MIM training blindly encourages the model to rewrite those texts rather than remove them, resulting in conflict with the STR target.
To address this conflict, we design a Background Modeling (BM) stream to focus on background content generation, which develops MIM in two aspects.
Firstly, we introduce text detection labels to prevent optimization from reconstructing text regions.
Secondly, we replace the random mask with a text region mask from another image, aligning the shape and size of recovery regions with the text regions.

Specifically, given an input image $I$ and the corresponding binary text region mask $M_{text}$ (1 means the text region). 
We first use $M_{text}$ to mask all text regions and get the background image $I_{bg}$.
Then we random sample a new text region mask $M_{rand}$ from the dataset to delete the non-text regions in $I_{bg}$ for MIM training.
After preprocessing by $M_{text}$ and $M_{rand}$, the masked image $I_m$ is sent into the BM stream $F_B$ and we force the output $F_B(I_m)$ to recover the non-text regions in $I_{bg}$.
The total optimization process can be formulated as follows,
\begin{equation}
    I_{bg}=I*(1-M_{text}),
\end{equation}
\begin{equation}
    I_m = I*(1-M_{text})*(1-M_{rand}),
\end{equation}
\begin{equation}
    L_{BM} = l(F_B(I_m)*(1-M_{text}), I_{bg}),
\end{equation}
where $l$ can be any recovery loss (e.g., L1 loss) and  $*$ denotes the Hadamard product. 
Compared with the general MIM method, the model in the BM stream always learns to predict non-text results during the whole pretraining stage, avoiding the learning of text textures. 
Thus, the model is biased to generate non-text backgrounds, resulting in three advantages: 
(1) Benefiting from the model not predicting text-like textures, the conflict between MIM and STR can be addressed, resulting in a consistent optimization target.
(2) By utilizing $M_{rand}$ collected from real-world text region masks, our BM stream enables the model to adapt to the shapes of text regions, enhancing its ability to remove texts.
(3) Since we also add masks on text regions using $M_{text}$, the BM predictions on $M_{text}$ can be seen as the coarse text removal labels for the TE stream. 

\subsection{Text Eraseing stream}
Although the BM stream can capture strong background generation knowledge for text removal, it still needs the guidance of text region mask $M_{text}$ to localize the regions for generation.
To achieve end-to-end text removal, we further propose a Text Eraseing (TE) stream to simultaneously learn text localization perception and text texture removal.

Concretely, to ensure the flexibility for various STR model structures, the TE stream just adopts the standard end-to-end STR process: the text image $I$ is directly fed into the TE stream $F_T$ to generate final STR predictions.
During optimization, we utilize the text region masks ($M_{text}$) to combine the background regions of the input image $I$ and the text regions of BM results ($F_B(I_m)$) as the pseudo label.
The loss is calculated as follows, 
\begin{equation}
    I_{pseudo} = I*(1-M_{text}) + F_B(I_m)*M_{text},
    \label{eq:L_psedo}
\end{equation}
\begin{equation}
    L_{TE} = l(F_T(I), I_{pseudo}).
    \label{eq:L_TI}
\end{equation}

By replacing all texts in the image with BM predictions as the label, we achieve end-to-end weakly supervised STR learning that leads the model to automatically overwrite non-text backgrounds on text regions.
With explicitly optimizing the STR target during pretraining, our TE stream further improves the learning efficiency on STD datasets, resulting in high-quality pretrained models.

\subsection{Prompt-based Multi-task Learning}
Since both streams utilize the same model but take different inputs and labels, we design a Prompt-based Multi-task Learning (PML) strategy to lead the model performing the required stream.
Concretely, we define two 1D vectors as prompts for each stream.
In the forward process, we first use the encoder to extract the image feature.
Then, the vector for the corresponding task will be duplicated and added to the feature map element-wise, providing a bias to distinguish different tasks.
Finally, the biased feature is sent to the decoder for prediction.
Thus, the PML does not need to modify the model structure and can be used in most of the end-to-end STR models seamlessly.
With PML, the two streams can be trained in parallel without introducing many additional parameters or computations.
As a result, the background generation knowledge learned from the BM stream can be exhaustively utilized for text removal.
This enhances the efficiency of our framework as well streamlines the pretraining process into one model and one stage.
Results in \cref{sec:experiments} confirm the effectiveness of this strategy in training models with multiple tasks in TMIM.

\subsection{Optimization}
For the pretraining stage, we combine $F_B(I_m)$ and $F_T(I)$ for joint optimization, both BM and TE streams use the same recovery loss function and the total loss is the sum of them.
Inspired by \cite{wang2023real}, the loss can be divided into the following parts: content loss, SSIM loss, and feature loss.
To enhance clarity, we denote both $F_B(I_m)$ and $F_T(I)$ as $O$.
And their label $I_{bg}$ and $I_{pseudo}$ are denoted as $Y$.

\noindent\textbf{Content loss.}
This loss assumes the prediction result to be the same as the label in both RGB (L1 loss) and frequency (FFL loss\cite{jiang2021focal}) domains:
\begin{equation}
    L_{C} = \lambda_1 L1(O,Y) + \lambda_2 FFL(O,Y),
    \label{eq:l_c}
\end{equation}
where both $\lambda_1$ and $\lambda_2$ are set as 15.

\noindent\textbf{SSIM loss.}
Based on \cite{wang2004image}, We further introduce SSIM as a training target to optimize the structure similarity between $O$ and its label.
\begin{equation}
    \begin{split}
        L_{SSIM} = 1-SSIM(O, Y).
    \end{split}
    \label{eq:l_c}
\end{equation}

\noindent\textbf{Feature loss.}
This loss is commonly used in image-to-image tasks to ensure the content of prediction and label have the same image style.
It uses a pretrained VGG net to extract feature maps and calculate their similarity as the optimization target.
We sum the GS loss and VGG loss defined in \cite{wang2023real} as our feature loss $L_{F}$ to optimize the style of $O$.

Overall, the combined pretraining loss for the BM and TE streams is defined as:
\begin{equation}
    L_{BM/TE} = L_{C}+\alpha L_{SSIM}+\beta L_{F},
    \label{eq:l_biti}
\end{equation}
where both $\alpha$ and $\beta$ are set to 1.
And the total loss is calculated by:
\begin{equation}
    L = L_{BM}+L_{TE}.
    \label{eq:l_total}
\end{equation}

Finally, as the finetuning stage also follows the TE stream, we simply use $L_{TE}$ as the finetuning loss.

\section{Experiments}
\label{sec:experiments}

\subsection{Datasets and Implementation Details}
\label{sec:dataset}
\label{sec:imp}
\noindent\textbf{Scene text detection datasets.}
We combine 8 public STD datasets for pretraining, including Total Text \cite{ch2017total}, ICDAR2015 \cite{karatzas2015icdar}, MLT19 \cite{nayef2019icdar2019}, ReCTS \cite{zhang2019icdar}, ArT \cite{chng2019icdar2019}, LSVT \cite{sun2019icdar}, COCO Text \cite{veit2016coco}, and Text OCR \cite{singh2021textocr}. We remove the overlapped test images in SCUT-EnsText \cite{liu2020erasenet} and finally get 111066 images.

\noindent\textbf{Scene text removal datasets.}
We utilize two commonly used STR datasets for finetuing and testing, including SCUT-EnsText \cite{liu2020erasenet}and SCUT-Syn \cite{zhang2019ensnet}.
Concretely, SCUT-EnsText is a real-world dataset. It contains 2749 images for training and 813 images for testing. 
SCUT-Syn is a synthetic dataset that contains 8000 images for training and 800 images for testing.

% \subsection{Implementation Details}
% \label{sec:imp}
\noindent\textbf{Model architectures.} We use Uformer \cite{wang2022uformer} as our baseline STR model.
Besides, to evaluate the generalization of our TMIM, we further introduce other advanced open-source STR architectures for evaluation, including Erasenet \cite{liu2020erasenet}, PERT \cite{wang2023real}, CTRNet \cite{liu2022don}, and FETNet \cite{lyu2023fetnet}.
Especially, for CTRNet, we remove the text perception module to avoid needing the additional STD model.

\noindent\textbf{Optimization details.}
In the training stage, we use the AdamW optimizer \cite{loshchilov2017decoupled} with the momentum of (0.9, 0.999) for optimization.
Following \cite{wang2022uformer}, the learning rate and weight decay are set to 2e-4 and 0.02.
For augmentation, we employ random flip and random jitter on color, brightness. 
We pretrain the model on the combined STD dataset for 10 epochs with the batch size of 8.
The pretrained models are finetuned on the target STR dataset for 200 epochs with the batch size of 16.
All images are resized to $512 \times 512$.
The experiments are employed with the PyTorch framework on 8 NVIDIA RTX 3090 GPUs.

% \subsection{Evaluation Metrics}
\noindent\textbf{Evaluation Metrics.}
Following the previous STR method \cite{liu2020erasenet}, we evaluate our framework with both Image-Eval and Detection-Eval metrics.
For Image-Eval metrics, we utilize PSNR, MSSIM, MSE, AGE, pEPs and pCEPs.
We follow \cite{peng2023viteraser} to multiply MSE with 100\%.
For Detection-Eval metrics, we use the pretrained CRAFT model \cite{baek2019character} as the text detector to test the precision (P), recall (R), and f-measure (F) on the erased images.
If not specified, we use SCUT-EnsText as the test dataset.

\begin{table}[!t]
    \caption{The influence of different modules in TMIM. Bold values denote the best results in each column. The baseline model is  Uformer-T.}
    \centering
    \begin{tabular}{l|c|c|c|c|c|c}
        \toprule
        Methods & PSNR$\uparrow$ & MSSIM$\uparrow$ & MSE$\downarrow$ & AGE$\downarrow$ & pEPs$\downarrow$ & pCEPs$\downarrow$ \\
        \midrule
        TE(baseline) & 34.07 & 97.10 & 0.12 & 2.04 & 0.0130 & 0.0085 \\
        \quad +MIM & 34.53 & 97.10 & 0.11 & 1.98 & 0.0121 & 0.0076 \\
        \midrule
        \quad +BM & 36.00 & \textbf{97.54} & \textbf{0.08} & 1.76 & 0.0095 & 0.0060 \\
        \quad +BM/PML(TMIM) & \textbf{36.18} & 97.48 & \textbf{0.08} & \textbf{1.75} & \textbf{0.0092} & \textbf{0.0056} \\
        \bottomrule
    \end{tabular}
    \label{tab:abl_module}
\end{table}

\begin{table}[!t]
\caption{The STR ability of pretrained models. "FS" denotes fully supervised training on the STR dataset. "WS" denotes weakly supervised pretraining on the STD dataset.}
\centering
\begin{tabular}{l|c|c|c|c|c}
    \toprule
    Methods & Strategy & PSNR$\uparrow$ & MSSIM$\uparrow$ & MSE$\downarrow$ & AGE$\downarrow$\\
    \midrule
    Uformer-T & FS & 34.07 & 97.10 & 0.12 & 2.04\\
    Uformer-T+TMIM & WS & \textbf{35.59(1.52$\uparrow$)} & \textbf{97.22(0.12$\uparrow$)} & \textbf{0.08(0.04$\downarrow$)} & \textbf{1.91(0.13$\downarrow$)}\\
    \midrule
    Uformer-S & FS & 34.81 & 97.36 & 0.11 & 1.90 \\
    Uformer-S+TMIM & WS & \textbf{36.05(1.24$\uparrow$)} & \textbf{97.26(0.10$\uparrow$)} & \textbf{0.08(0.03$\downarrow$)} & \textbf{1.85(0.05$\downarrow$)}\\
    \midrule
    Uformer-B & FS & 35.17 & 97.35 & 0.11 & 1.90\\
    Uformer-B+TMIM & WS & \textbf{36.62(1.45$\uparrow$)} & \textbf{97.36(0.01$\uparrow$)} & \textbf{0.06(0.05$\downarrow$)} & \textbf{1.71(0.19$\downarrow$)} \\
    \midrule
    \bottomrule
\end{tabular}
\label{tab:abl_pretrain}
\end{table}

\subsection{Ablation Study}
\label{sec:abl}

\noindent\textbf{The effectiveness of each module in TMIM.}
As shown in \cref{tab:abl_module}, the Uformer-T without pretraining can be seen as the TMIM with only TE stream, resulting in 34.07 PSNR. 
And with general MIM pretraining, it obtains 34.53 PSNR.
After introducing our BM, we get an obvious improvement of 1.47 PSNR. 
This benefits from the explicit STR learning and the pseudo labels guiding.
With the PML, the model finally achieves 36.18 PSNR which surpasses the baseline with and without pretraining by 2.11 and 1.65.
Overall, the improvements demonstrate that:
(1) Our method can explore more valuable knowledge from existing STD data to better alleviate the limitation of the small-scale STR dataset.
(2) Introducing text localization information to learn non-text content generation and generating pseudo labels for explicit STR learning is more suitable than implicit feature learning in MIM.
(3) The task prompt can well balance the two streams in one model for further improvement. 

\noindent\textbf{The STR ability of pretrained model.}
As the pretrained models in our TMIM also have end-to-end STR ability, we list their performance in \cref{tab:abl_pretrain}.
We compare our weakly supervised pretrained models trained on STD datasets with the fully supervised models trained on STR datasets.
\cref{tab:abl_pretrain} shows that our pretrained Uformer surpasses the fully supervised baseline on all 3 sizes.
Especially, our Uformer-B+TMIM obtains 36.62 PSNR, which also outperforms all non-finetuning STR models in \cref{tab:sota_ens} (36.62 vs 36.32).
This proves that our TMIM can explicitly learn valuable STR knowledge from STD datasets without finetuning, providing new insight for label-constrained STR learning.

\begin{table}[!t]
    \caption{The generalization with different STR architectures.}
    \centering
    \begin{tabular}{l|c|c|c|c}
        \toprule
        Methods & PSNR$\uparrow$ & MSSIM$\uparrow$ & MSE$\downarrow$ & AGE$\downarrow$\\
        \midrule
        Erasenet & 32.30 & 95.42 & 0.15 & 3.02\\
        \quad+TMIM & \textbf{33.63(1.33$\uparrow$)} & \textbf{97.20(1.78$\uparrow$)} & \textbf{0.11(0.04$\downarrow$)} & \textbf{2.45(0.57$\downarrow$)}\\
        \midrule
        PERT & 33.62 & 97.00 & 0.13 & 2.19 \\
        \quad+TMIM & \textbf{35.21(1.59$\uparrow$)} & \textbf{97.15(0.15$\uparrow$)} & \textbf{0.09(0.04$\downarrow$)} & \textbf{1.88(0.31$\downarrow$)}\\
        \midrule
        CTRNet & 35.20 & 97.36 & 0.09 & 2.20\\
        \quad+TMIM & \textbf{35.54(0.34$\uparrow$)} & \textbf{97.48(0.12$\uparrow$)} & \textbf{0.07(0.02$\downarrow$)} & \textbf{2.15(0.05$\downarrow$)} \\
        \midrule
        FETNet & 34.53 & 97.01 & 0.13 & \textbf{1.75}\\
        \quad+TMIM & \textbf{35.16(0.63$\uparrow$)} & \textbf{97.22(0.21$\uparrow$)} & \textbf{0.11(0.02$\downarrow$)} & 1.96(-0.21$\downarrow$) \\
        \midrule
        Uformer-B & 35.17 & 97.35 & 0.11 & 1.90\\
        \quad+TMIM & \textbf{37.35(2.18$\uparrow$)} & \textbf{97.78(0.43$\uparrow$)} & \textbf{0.05(0.06$\downarrow$)} & \textbf{1.61(0.29$\downarrow$)} \\
        \midrule
        \bottomrule
    \end{tabular}
    \label{tab:abl_arch}
\end{table}

\begin{table}[!t]
\caption{Comparison with other pretraining methods.}
\centering
\begin{tabular}{l|c|c|c|c|c|c|c}
    \toprule
    Method & Pretrain & PSNR$\uparrow$ & MSSIM$\uparrow$ & MSE$\downarrow$ & AGE $\downarrow$ & pEPs $\downarrow$ & pCEPs $\downarrow$\\
    \midrule
    \multirow{4}{*}{PERT} & $\times$ & 33.62 & 97.00 & 0.13 & 2.19 & 0.0135 & 0.0088 \\
    & MIM & 33.81 & 97.10 & 0.1203 & 2.07 & 0.0146 & 0.0094\\
    & SegMIM & 34.92 & 97.13 & 0.0963 & 1.89 & 0.0121 & 0.0076 \\
    & TMIM(ours) & \textbf{35.21} & \textbf{97.15} & \textbf{0.0942} & \textbf{1.88} & \textbf{0.0119} & \textbf{0.0074}\\
    \midrule
    \multirow{4}{*}{Uformer-S} & $\times$ & 34.81 & 97.36 & 0.1128 & 1.90 & 0.0119 & 0.0077 \\
    & MIM & 36.03 & 97.45 & 0.0832 & 1.74 & 0.0093 & 0.0057\\
    & SegMIM & 36.68 & 97.56 & 0.0684 & 1.69 & 0.0086 & 0.0053\\
    & TMIM(ours) & \textbf{36.87} & \textbf{97.74} & \textbf{0.0614} & \textbf{1.62} & \textbf{0.0078} & \textbf{0.0048} \\
    \bottomrule
\end{tabular}
\label{tab:sota_pretrain}
\end{table}

\noindent\textbf{The generalization of TMIM.}
As a model-agnostic framework, to evaluate the generalization of our TMIM, we test our TMIM on 5 advanced STR models including both CNN and Transformer architectures: Erasenet \cite{liu2020erasenet}, PERT \cite{wang2023real}, CTRNet \cite{liu2022don}, FETNet \cite{lyu2023fetnet}, and Uformer \cite{wang2022uformer}. 
As illustrated in \cref{tab:abl_arch}, our TMIM can enhance the overall performance on multiple STR architectures.
All results on PSNR, MSSIM, and MSE metrics are improved.
This demonstrates that our pretraining method is a general framework for further improving the performance of different STR model architectures.

\noindent\textbf{Comparison with other pretraining methods.}
In \cref{tab:sota_pretrain}, we reimplement other pretraining methods for comparison, including MIM and SegMIM.
We choose a CNN-based (PERT \cite{wang2023real}) and a Transofmer-based (Uformer-S \cite{wang2022uformer}) architecture as the baseline model.
For a fair comparison, we also pretrain the baseline model with the compared methods on our 111k STD images and finetune them on the SCUT-EnsText dataset.
It is shown that our TMIM surpasses other pretrain methods on all metrics.
These results demonstrate that our method is a more suitable framework for the STR task, which can exhaustively explore the upper bound
of multiple models.

\begin{table}[t]
    \caption{Comparison with SOTA STR models on SCUT-EnsText. Bold and underlined values denote the 1st and 2nd results in each column. $\dagger$ denotes our reimplement. $\ddagger$ denotes pretrained on STD datasets.}
    \centering
    \resizebox{\textwidth}{!}  
    {
    \begin{tabular}{l|c c c c c c|c c c | c} 
        \toprule
        \multirow{2}{*}{Methods} & \multicolumn{6}{c|}{Image-Eval} & \multicolumn{3}{c|}{Detection-Eval} & \multirow{2}{*}{Param} \\
        \cline{2-10}
        & PSNR$\uparrow$ & MSSIM$\uparrow$ & MSE$\downarrow$ & AGE$\downarrow$ & pEPs$\downarrow$ & pCEPs$\downarrow$ & R$\downarrow$ & P$\downarrow$ & F$\downarrow$
        \\
        \midrule
        Original & - & - & - & - & - & - & 69.5 & 79.4 & 74.1  & - \\
        EraseNet\cite{liu2020erasenet} & 32.30 & 95.42 & 0.15 & 3.02 & 0.0160 & 0.0090 & 4.6 & 53.2 & 8.5  & 17.8\\
        PERT\cite{wang2023real} & 33.62 & 97.00 & 0.13 & 2.19 & 0.0135 & 0.0088 & 4.1 & 50.5 & 7.6  & 14\\
        Tang et al.\cite{tang2021stroke} & 35.54 & 96.24 & 0.09 & - & - & - & 3.6 & - & -  & 30.7\\
        PSSTRNet\cite{lyu2022psstrnet} & 34.65 & 96.75 & 0.14 & 1.72 & 0.0135 & 0.0074 & 5.1 & 47.7 & 9.3  & 4.9\\
        CTRNet\cite{liu2022don} & 35.20 & 97.36 & 0.09 & 2.20 & 0.0106 & 0.0068 & 1.4 & 38.4 & 2.7  & 159.8\\
        MBE\cite{hou2022multi} & 35.03 & 97.31 & - & 2.06 & 0.0128 & 0.0088 & - & - & -  & -\\
        SAEN\cite{du2023modeling} & 34.75 & 96.53 & 0.07 & 1.98 & 0.0125 & 0.0073 & - & - & -  & 19.8\\
        FETNet\cite{lyu2023fetnet} & 34.53 & 97.01 & 0.13 & 1.75 & 0.0137 & 0.0080 & 5.8 & 51.3 & 10.5  & 8.5\\
        Uformer-B$\dagger$\cite{wang2022uformer} & 35.17 & 97.35 & 0.11 & 1.90 & 0.0107 & 0.0069 & 2.9 & 51.6 & 5.5 & 50.88\\
        ViTEraser\cite{peng2023viteraser} & 36.32 & 97.51 & 0.0565 & 1.86 & 0.0074 & \underline{0.0041} & 0.635 & 37.8 & 1.248  & 191.9\\
        ViTEraser$\ddagger$\cite{peng2023viteraser} & \underline{37.11} & 97.61 & \underline{0.0474} & 1.70 & \underline{0.0066} & \textbf{0.0035} & \textbf{0.389} & \underline{29.7} & \textbf{0.768}  & 191.9\\
        \midrule
        Uformer-T+TMIM$\ddagger$ & 36.18 & 97.48 & 0.0813 & 1.75 & 0.0092 & 0.0056 & 1.457 & 45.2 & 2.824 & 5.29\\
        Uformer-S+TMIM$\ddagger$ & 36.87 & \underline{97.74} & 0.0614 & \underline{1.62} & 0.0078 & 0.0048 & 0.595 & \textbf{12.4} & 1.113 & 20.75\\
        Uformer-B+TMIM$\ddagger$ & \textbf{37.35} & \textbf{97.78} & \textbf{0.0470} & \textbf{1.61} & \textbf{0.0065} & \underline{0.0041} & \underline{0.533} & 34.7 & \underline{1.051} & 50.88\\
        \bottomrule
    \end{tabular}
    }
    \label{tab:sota_ens}
\end{table}

\begin{table}[t]
\centering
\caption{Comparison with SOTA STR models on SCUT-Syn. Bold values denote the best results in each column. $\dagger$ denotes our reimplement. $\ddagger$ denotes pretrained on STD datasets.}
{
    \begin{tabular}{l|c c c c c c}
        \toprule
        Methods & PSNR$\uparrow$ & MSSIM$\uparrow$ & MSE$\downarrow$ & AGE$\downarrow$ & pEPs$\downarrow$ & pCEPs$\downarrow$ \\
        \midrule
        EraseNet\cite{liu2020erasenet} & 38.32 & 97.67 & 0.02 & 1.60 & 0.0048 & 0.0004\\
        PERT\cite{wang2023real} & 39.40 & 97.87 & 0.02 & 1.41 & 0.0046 & 0.0007\\
        Tang et al.\cite{tang2021stroke} & 38.60 & 97.55 & 0.02 & - & - & -\\
        MBE\cite{hou2022multi} & 43.85 & 98.64 & - & \textbf{0.94} & 0.0013 & 0.00004\\
        SAEN\cite{du2023modeling} & 38.63 & 98.27 & 0.03 & 1.39 & 0.0043 & 0.0004\\
        FETNet\cite{lyu2023fetnet} & 39.14 & 97.97 & 0.02 & 1.26 & 0.0046 & 0.0008\\
        Uformer-B$\dagger$\cite{wang2022uformer} & 40.60 & 98.37 & 0.0156 & 1.29 & 0.0032 & 0.0004\\
        ViTEraser\cite{peng2023viteraser} & 42.53 & 98.45 & 0.0102 & 1.19 & 0.0018 & 0.000016\\
        ViTEraser$\ddagger$\cite{peng2023viteraser} & 42.97 & 98.55 & 0.0092 & 1.11 & 0.0015 & 0.000011\\
        \midrule
        Uformer-T+TMIM$\ddagger$ & 41.81 & 98.62 & 0.0125 & 1.16 & 0.0023 & 0.000149 \\
        Uformer-S+TMIM$\ddagger$ & 42.59 & 98.73 & 0.0111 & 1.11 & 0.0019 & 0.000087\\
        Uformer-B+TMIM$\ddagger$ & \textbf{44.24} & \textbf{98.86} & \textbf{0.0084} & 0.95 & \textbf{0.0013} & \textbf{0.000011} \\
        \bottomrule
    \end{tabular}
}
\label{tab:sota_syn}
\end{table}

\begin{figure}[!t]
\centering
    \includegraphics[width=0.95\linewidth]{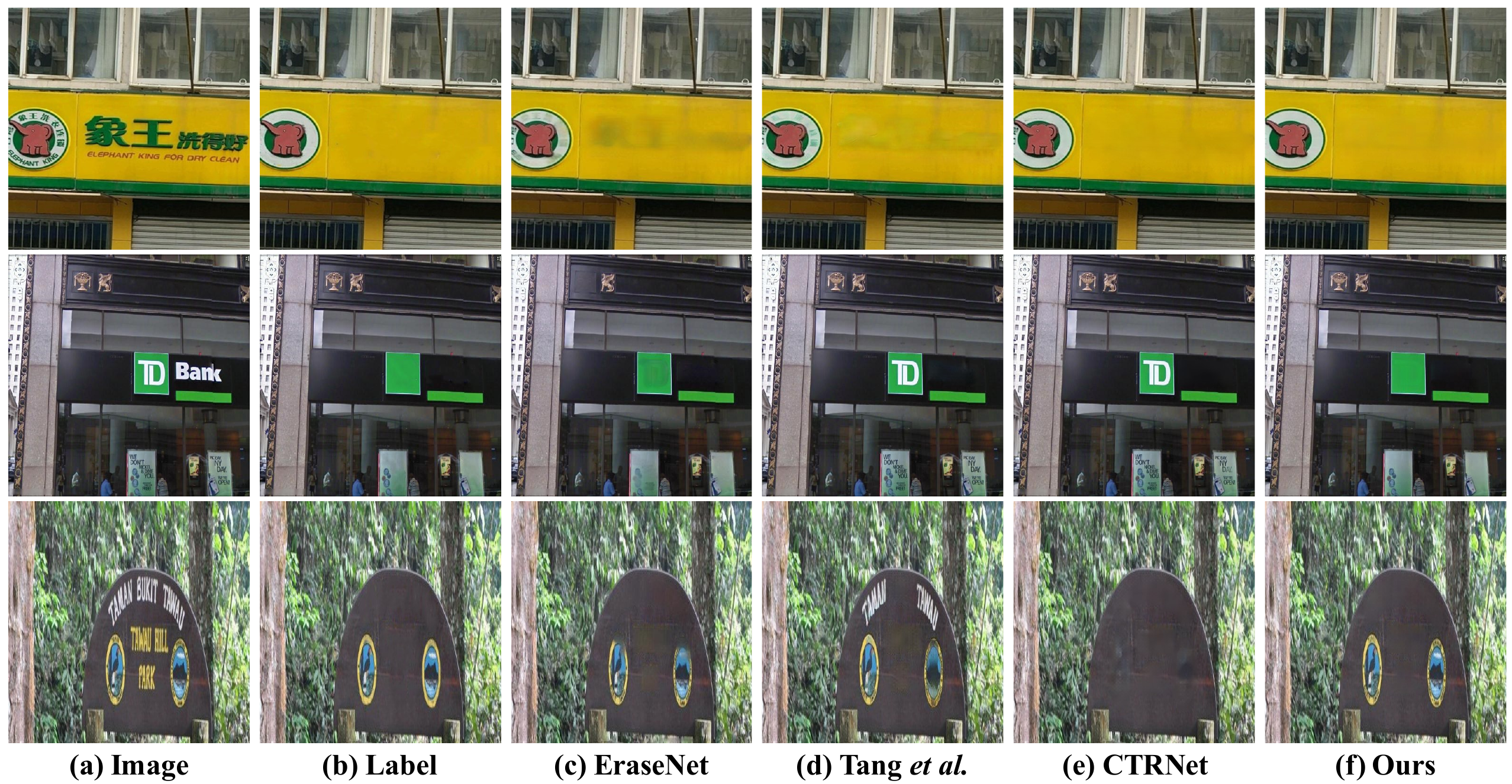}
    \caption{The qualitative results on SCUT-EnsText compared with Erasenet\cite{liu2020erasenet}, Tang et al.\cite{tang2021stroke}, and CTRNet\cite{liu2022don}.}
    \label{fig:img_vis}
\end{figure} 

\subsection{Comparison with State-of-the-Art Methods}

In \cref{tab:sota_ens}, we first apply our TMIM on Uformer and compare it with other advanced STR methods on the real SCUT-EnsText dataset.
It can be seen that our Uformer-B+TMIM outperforms other methods, achieving state-of-the-art PSNR$\uparrow$ (37.35 vs 37.11), SSIM$\uparrow$ (97.78 vs 97.61), MSE$\downarrow$ (0.0470 vs 0.0474), and pEPS$\downarrow$ (0.0065 vs 0.0066).
Especially, compared with the recently proposed ViTEraser \cite{peng2023viteraser} which also uses large-scale STD data for pretraining, our Uformer-B+TMIM gets better results on most image-eval metrics with a lighter pretraining process (10 epochs with 111k images vs 100 epochs with 98K images) and model parameters (50.88M vs 191.9M).
For detection-eval metrics, \cite{peng2023viteraser} performs better while it is sensitive to the detection model, which may be jitter under other detectors.
This verifies that our weakly supervised STR learning can capture more STR knowledge efficiently.
Besides, we illustrated some representative results in \cref{fig:img_vis}. 

In \cref{tab:sota_syn}, we also test our methods on the synthetic SCUT-Syn dataset.
Results show that our method reaches state-of-the-art performance except for MBE \cite{hou2022multi} in the AGE metric, which combines multiple networks for prediction. And the overall results again verify the superiority of our method.

\begin{table}[!t]
\caption{Comparison with inpainting methods. $\dagger$ denotes using the ground truth of text region masks for testing.}
\centering
\begin{tabular}{l|c|c|c|c}
    \toprule
    Methods & PSNR$\uparrow$ & MSSIM$\uparrow$ & MSE$\downarrow$ & AGE$\downarrow$ \\ 
    \midrule
    CTSDG$\dagger$\cite{guo2021image} & 33.10 & 95.55 & 0.14 & - \\ 
    SPL$\dagger$\cite{zhang2021context} & 35.41 & 97.39 & 0.07 & - \\
    LaMa$\dagger$\cite{suvorov2022resolution} & 35.25 & 97.36 & 0.07 &1.97 \\ 
    MAT$\dagger$\cite{li2022mat} & 34.22 & 97.13 & 0.09 & 2.19 \\ 
    Uformer-B+TMIM(ours) & \textbf{37.35} & \textbf{97.78}  & \textbf{0.05} & \textbf{1.61} \\
    \bottomrule
\end{tabular}
\label{tab:abl_inpaint}
\end{table}

\begin{figure*}[!t]
\centering
    \includegraphics[width=0.8\textwidth]{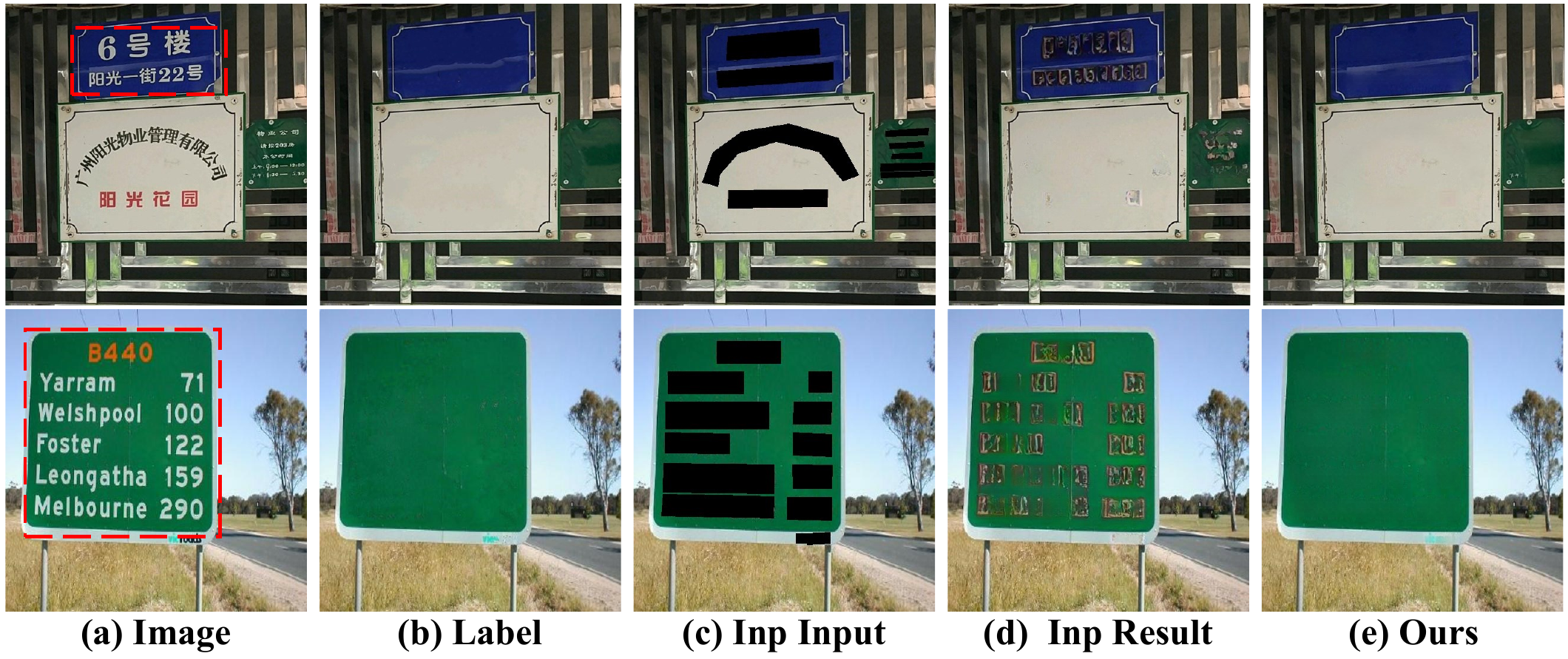}
    \caption{The qualitative results compared with the inpainting method Lama \cite{suvorov2022resolution}. (c) is the input for inpainting method masked by the text detection ground truth. (d) is the result of LaMa \cite{suvorov2022resolution}.}
    \label{fig:img_inp}
\end{figure*}

\subsection{Discussion}
\label{sec:discussion}

\noindent\textbf{Comparison with inpainting methods.}
Existing inpainting methods also can be used for text removal when providing additional text detection results on the test set.
However, similar to MIM, the training target of the inpainting task is also to reconstruct the original contents, which results in unexpected text-like textures rather than removing the texts on scenes where text frequently appears  (as illustrated in \cref{fig:img_inp}).
\cite{liu2020erasenet} also shows that inpainting models perform poorly on recovering text regions as there is a domain gap between STR and inpainting tasks.
Moreover, we introduce the ground truth of the text region for inpainting models to test the STR ability in \cref{tab:abl_inpaint}.
The results of CTSDG\cite{guo2021image} and SPL\cite{zhang2021context} are collected from \cite{liu2022don}.
And we further reimplement LaMa\cite{suvorov2022resolution} and MAT\cite{li2022mat} using the STD and STR dataset in \cref{sec:dataset}.
It is obvious that our TMIM obtains better performance without the need for text detection ground truth (37.35 vs 35.41 PSNR), verifying our method is more suitable for the STR task.

\noindent\textbf{Comparison with Detection-based STR methods.}
In contrast to using the STD data for pretraining, there are also many two stage STR methods \cite{tang2021stroke,liu2022don,lee2022surprisingly,feng2024deeperaser} that introduce additional pretrained STD models to predict text region masks as the auxiliary information.
The main difference between our TMIM and them lies in: 
(1) We only need text region masks during the training stage. In the inference stage, both the pretrained and finetuned models in our TMIM can directly remove the texts in a simple one-stage process without using text detection results.
(2) As listed in \cref{tab:abl_detection}, our method without text detection results also outperforms these two stage STR methods (37.35 vs 35.84 PSNR).
After following \cite{lee2022surprisingly} to only focus on the results in text regions, our model finally achieves state-of-the-art 43.88 PSNR, demonstrating our method is more efficient than existing detection-based STR methods.

\begin{table}[!t]
\caption{Comparison with detection-based STR methods. $\dagger$ denotes using the text region masks from \cite{baek2019character}. $\ddagger$ denotes only concern the results on text region follow \cite{lee2022surprisingly}.}
\centering
\begin{tabular}{l|c|c|c|c}
    \toprule
    Methods & PSNR$\uparrow$ & MSSIM$\uparrow$ & MSE$\downarrow$ & AGE$\downarrow$ \\ 
    \midrule
    Tang et al.$\dagger$\cite{tang2021stroke} & 35.54 & 97.00 & 0.09 & - \\ 
    CTRNet$\dagger$\cite{liu2022don} & 35.20 & 97.36 & 0.09 & 2.20 \\ 
    DeepEraser$\dagger$\cite{feng2024deeperaser} & 35.84 & 97.48 & 0.08 & 1.71 \\ 
    Uformer-B+TMIM(ours) & \textbf{37.35} & \textbf{97.78}  & \textbf{0.05} & \textbf{1.61} \\ 
    \midrule
    CTRNet$\ddagger$\cite{liu2022don} & 37.20 & 97.66 & 0.07 & - \\ 
    GaRNet$\ddagger$\cite{lee2022surprisingly} & 41.37 & 98.46 & - & 0.59 \\ 
    DeepEraser$\ddagger$\cite{feng2024deeperaser} & 42.47 & 98.65 & - & 0.59 \\ 
    % \midrule
    Uformer-B+TMIM$\ddagger$(ours) & \textbf{43.88} & \textbf{98.88}  & \textbf{0.04} & \textbf{0.54} \\ 
    \bottomrule
\end{tabular}
\label{tab:abl_detection}
\end{table}

\section{Conclusions}
\label{sec:conclusions}
In this paper, we propose a weakly supervised STR pretraining framework to address the limitation of high-cost small-scale STR datasets, named Text-aware Masked Image Modeling (TMIM).
The TMIM introduces low-cost text localization for pretraining, which enables end-to-end STR training under weak supervision.
We build a BM stream to learn background generation rules from non-text regions and predict pseudo STR labels on text regions.
Meanwhile, we design a TE stream to learn text removal using pseudo labels.
Experiments show that our pretrained model already obtains impressive performance and finally achieves state-of-the-art performance after finetuning.
In the future, we will explore a more efficient STR learning method under low-cost text detection labeling, further enhancing the final STR performance.

~\\
\noindent\textbf{Acknowledgements.}
This work is supported by the National Key Research and Development Program of China (2022YFB3104700), the National Nature Science Foundation of China (62121002, U23B2028, 62102384).
We acknowledge the support of GPU cluster built by MCC Lab of Information Science and Technology Institution, USTC.
We also thank the USTC supercomputing center for providing computational resources for this project.

% ---- Bibliography ----
%
% BibTeX users should specify bibliography style 'splncs04'.
% References will then be sorted and formatted in the correct style.
%
\bibliographystyle{splncs04}
\bibliography{main}
\end{document}